\documentclass{article} 
\usepackage{iclr2027_conference,times}

\usepackage{amsmath,amsfonts,bm}

\def\eqref#1{equation~\ref{#1}}

\def\1{\bm{1}}

\DeclareMathAlphabet{\mathsfit}{\encodingdefault}{\sfdefault}{m}{sl}
\SetMathAlphabet{\mathsfit}{bold}{\encodingdefault}{\sfdefault}{bx}{n}

\usepackage{multirow}
\usepackage{hyperref}
\usepackage{url}
\usepackage{graphicx}
\usepackage[table]{xcolor}
\usepackage{booktabs}
\usepackage{makecell}
\usepackage{caption}
\usepackage{amssymb}
\usepackage{pifont}
\newcommand{\cmark}{\ding{51}}%
\newcommand{\xmark}{\ding{55}}%
\definecolor{citecolor}{RGB}{66,168,235}
\definecolor{linkcolor}{RGB}{255,0,0}
\definecolor{oursrow}{RGB}{230,240,252}
\newcommand{\oursrow}{\rowcolor{oursrow}}
\newcommand{\ourscell}{\cellcolor{oursrow}}
\newcommand{\grouprow}[2]{\multicolumn{#1}{l}{\textit{#2}}}

\hypersetup{colorlinks=true,citecolor=citecolor,linkcolor=linkcolor}

\title{Joint Latent Reasoning and Observation for Agentic Video Understanding}

\author{%
  \parbox[t]{\dimexpr\textwidth-2\tabcolsep\relax}{%
    \raggedright\normalfont
    \textbf{Junzhang~Liu}$^{1}$, 
    \textbf{Hao~Yuan}$^{2}$, 
    \textbf{Yiming~Wu}$^{3}$, 
    \textbf{Qi~Zhang}$^{4}$, 
    \textbf{Long~Chen}$^{1\dagger}$, 
    \textbf{Zhecan~Wang}$^{2}$\thanks{Corresponding author}\\[0.5em]
    \small
    $^{1}$Hong Kong University of Science and Technology
    $^{2}$Shanghai Artificial Intelligence Laboratory\\
    $^{3}$The University of Hong Kong \quad
    $^{4}$Fudan University 
  }%
}
\begin{document}

\maketitle
\begin{abstract}
Effective video agents must learn not only how to answer, but also where to look next. Transferring this capability to smaller models is challenging because each observation serves two purposes: supporting an answer and guiding the search for further evidence. Textual descriptions may capture what is currently visible while omitting cues needed for subsequent acquisition. We introduce \textbf{Latent Agentic Reasoning for Video Analysis (LARVA)}, a framework that distills multi-round video agents through jointly learned latent reasoning and visual observations. Rather than communicating through captions, an observer passes compact continuous representations to a reasoner that selects the next acquisition action or produces the final answer. Crucially, action and answer supervision propagate through this interface, training the observer to communicate information useful for downstream decisions. Observation reconstruction and reasoning-distribution alignment provide auxiliary supervision from textual teacher trajectories. Across five video-understanding benchmarks, LARVA achieves strong performance with 4B and 7B backbones. On VideoHolmes, Video-MME, and LVBench, the 7B model outperforms a textual CoT student trained on the same teacher trajectories by 5.1, 2.5, and 3.7 percentage points, respectively. These results highlight the value of jointly learning how agents represent evidence and act on it when distilling video-understanding capabilities.

\end{abstract}

\vspace{-15pt}
\section{Introduction}
\vspace{-8pt}

A video agent succeeds through a sequence of interdependent decisions: observe part of a video, reason about the available evidence, decide where to look next, and incorporate the new observation before answering. Recent agentic video systems implement this loop with tools that selectively inspect temporal regions \citep{lenswalk, videoseek, liu-etal-2026-longvideoagent}. Each observation therefore affects both the interpretation of current evidence and the acquisition of future evidence. A missed temporal cue can redirect the search or cause premature stopping, leaving later reasoning without the evidence it needs. Distilling this capability into a smaller model requires preserving the coupling between observation, reasoning, and action throughout the interaction.

The representations passed around this loop determine what information can influence each decision. Many video agents express intermediate reasoning as textual chain-of-thought (CoT)~\citep{wei2022chain}; modular systems also communicate observations through descriptions or query-specific responses \citep{videoseek, liu-etal-2026-longvideoagent}. This repeatedly requires selecting which visual and reasoning details to verbalize. A description sufficient to identify an event may omit the temporal context needed to choose the next interval, leaving subsequent reasoning without direct access to that cue. Passing selected frames directly to the reasoner avoids textualization but adds visual tokens to its accumulating context \citep{EVA, videomind}. We ask: \emph{how can latent representations improve the complete observation--reasoning--action loop, from using current evidence to acquiring the evidence needed next?}

Continuous latent representations provide a promising foundation. DyLaR combines latent perception and reasoning for video question answering, MACF communicates compact visual representations among video agents, and LatentRAG jointly learns latent reasoning and retrieval for iterative search \citep{dylar, macf, latentrag}. Our focus is jointly distilling latent reasoning and visual observations within a closed-loop video agent. Latent observations condition reasoning about the next acquisition; that acquisition determines the next visual input, whose latent representation feeds back into subsequent reasoning. Their value must therefore be assessed through both the decisions made along the trajectory and the final answer.

\begin{figure*}[t]
    \centering
    \includegraphics[width=0.87\linewidth]{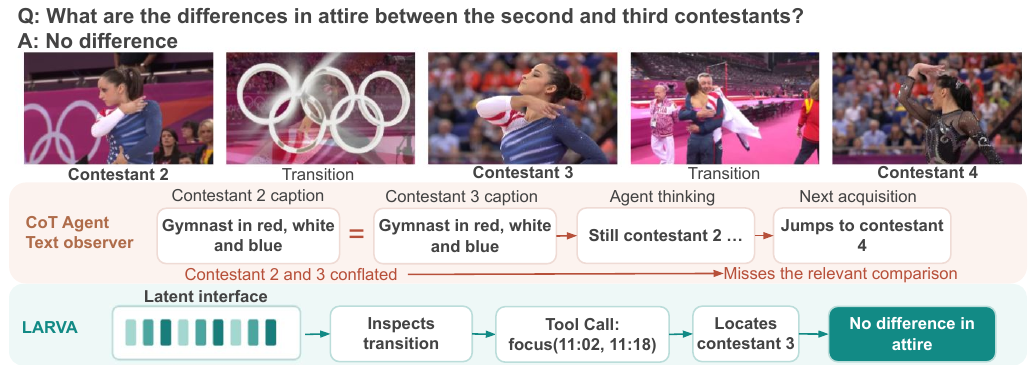}
    \caption{When asked to compare the contestants’ attire, the CoT agent receives identical captions (“Gymnast in red, white, and blue”), confuses their identities, and skips ahead to the fourth contestant. LARVA samples around the transitions and correctly identifies no difference in attire.}
    \label{fig:teaser}
    \vspace{-2em}
\end{figure*}

Learning this loop from textual teacher trajectories presents two challenges. First, the trajectories supervise reasoning traces, observations, and actions, but provide no direct targets for the latent states that replace the textual intermediates. Second, a representation useful in isolation may be ineffective in the loop: an observation latent may reconstruct a teacher description yet fail to support the reasoner's next decision. Our key design principle is to \emph{jointly shape latent observations and reasoning by the downstream actions and answers they support}. The continuous interface makes it possible to train how evidence is encoded together with how it is used, while textual supervision anchors the learned states to the teacher trajectory.

We introduce \textbf{Latent Agentic Reasoning for Video Analysis (LARVA)}, which implements this principle through a jointly trained reasoner--observer architecture. At each round, the reasoner uses accumulated observations to compute latent reasoning states and select an explicit tool call or final answer. The observer encodes the requested frames into latent tokens that directly condition the next reasoning step. This closes the feedback loop between representing evidence and selecting further evidence, without decoding intermediate reasoning or observations into text. Each reasoning and observation block contains eight latent tokens in our main configuration, while raw visual processing remains in the observer.

To train LARVA, we replay teacher acquisition actions while replacing textual intermediates with generated latent states. Later action and answer losses propagate through the latent observation interface, so the observer receives supervision from how the reasoner uses its output. Observation reconstruction anchors visual content, while CoT-distribution alignment trains the combined latent context to induce next-reasoning predictions similar to those induced by teacher text in a frozen reference model. These objectives connect visual encoding, intermediate reasoning, and explicit decisions during training. The teacher fixes acquisitions during replay; at inference, the learned representations guide the student's own acquisition loop.

Our contributions advance both the design and learning of the complete video-agent loop:
\begin{itemize}
    \item \textbf{A jointly learned latent observation--reasoning--action loop.}
    We introduce LARVA, which makes the representation of acquired visual evidence a trainable part of the acquisition policy. Latent observations feed directly into latent reasoning, which selects the next explicit action and thereby determines the evidence available in subsequent rounds. This formulation couples learning \emph{how to represent evidence}, \emph{how to reason over it}, and \emph{where to look next} within a single agentic framework.

    \item \textbf{Learning visual communication from downstream decisions.}
    We develop a trajectory-distillation objective that trains what the observer communicates through the actions and answers its output supports. Action and answer supervision jointly shapes the observer and reasoner, while reconstruction and CoT-distribution alignment transfer textual supervision to their latent states. Joint training improves the latent/latent configuration by 5.3--7.2 percentage points across three diagnostic benchmarks, taking it from below the textual baseline to the strongest evaluated representation configuration (Table~\ref{tab:repr_ablation}).

    \item \textbf{Stronger video understanding with acquisition gains across answerers.}
    LARVA improves over CoT students distilled from the same teacher trajectories on all five benchmarks with both 4B and 7B backbones, gaining 5.1, 2.5, and 3.7 points on VideoHolmes, Video-MME, and LVBench with the 7B model (Table~\ref{tab:main_results}). Its acquisition trajectories also improve accuracy with each of three fixed terminal answerers across all three diagnostic benchmarks; a fixed 27B visual answerer gains 1.7--4.2 points (Table~\ref{tab:evidence_quality}). These results connect stronger final answers to improvements in the evidence-acquisition process itself.
\end{itemize}

\vspace{-8pt}
\section{Related Work}
\vspace{-8pt}
\subsection{Agentic Video Understanding}
Because only a small fraction of frames in a long video is typically relevant to a query, recent work replaces exhaustive encoding with agents that iteratively decide what evidence to acquire. VideoSeek follows a think--act--observe loop with multi-granularity observation tools~\citep{videoseek}, and EVA trains a single multimodal large language model (MLLM) with supervised and reinforcement learning to plan what to watch, acquire frames, and reflect on the resulting evidence~\citep{EVA}. These systems differ in how reasoning and perception are coupled. Unified agents such as EVA interpret the acquired frames directly within the reasoning model. Reasoner--observer designs such as LensWalk and VideoSeek instead let a language-model reasoner choose the temporal range and sampling strategy, while a vision-language observer returns evidence~\citep{lenswalk, videoseek}. Other systems decompose the task further: LongVideoAgent coordinates grounding and vision agents~\citep{liu-etal-2026-longvideoagent}, VideoExplorer places a text-only reasoning model above a multimodal toolkit~\citep{yuan2025thinkvideosagenticlongvideo}, and VideoARM accumulates tool outputs in a hierarchical memory~\citep{Yin_2026_CVPR}. Many of these systems are training-free.

Across these designs, reasoning and perception communicate through explicit interfaces. Intermediate reasoning is generated as text, and modular observers typically convert visual evidence into textual descriptions for the reasoner~\citep{liu-etal-2026-longvideoagent, lenswalk}. Language is interpretable and convenient for combining independently developed models, but it requires sequential verbalization and projects high-dimensional visual evidence into a discrete representation before that evidence can influence subsequent decisions. LARVA retains the reasoner--observer decomposition but learns a continuous interface between the two roles.

\subsection{Latent Reasoning and Communication}
Latent reasoning replaces decoded reasoning steps with continuous hidden states. Coconut feeds the last hidden state back as the next input embedding and progressively replaces textual CoT steps through a curriculum~\citep{hao2024training}, while CODI distills explicit CoT into a compact latent sequence by aligning teacher and student hidden states~\citep{shen-etal-2025-codi}. In the visual domain, LaViT supervises latent visual thoughts with a teacher's visual semantics and attention patterns~\citep{wu2026lavitaligninglatentvisual}, and DyLaR combines separate perception and reasoning latents for video question answering over a fixed set of sampled frames~\citep{dylar}. Continuous representations have also been used for communication and interaction. MACF trains local video agents to pass compact latents to a coordinator~\citep{macf}; ALAR learns latent reasoning from action supervision and switches adaptively between latent and explicit reasoning across tool-use turns~\citep{jung2026adaptivelatentagenticreasoning}; and LatentRAG jointly trains a large language model (LLM) and a retriever for iterative search with latent reasoning and queries~\citep{latentrag}.
LARVA differs from them by jointly distilling latent reasoning and latent visual observations within a closed-loop video agent, so that action and answer losses shape the observation interface and the reasoning states.
\vspace{-8pt}
\section{Method}
\vspace{-8pt}
\label{sec:method}

\begin{figure*}[t]
    \centering
    \includegraphics[width=0.87\linewidth]{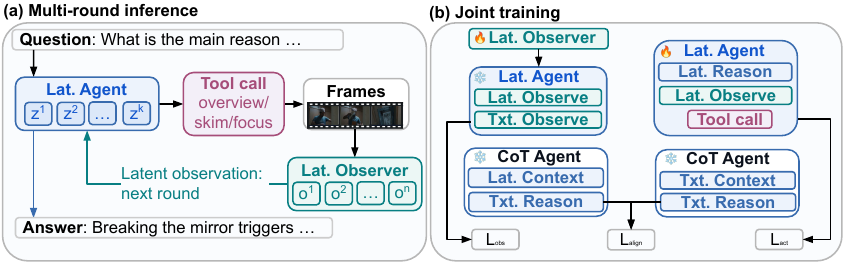}
    \caption{\textbf{Overview of LARVA. (a)} During multi-round inference, the latent agent selects video acquisitions and integrates compact latent observations to guide subsequent tool calls and produce an answer. \textbf{(b)} Joint training combines observation reconstruction \(\mathcal{L}_{\mathrm{obs}}\), CoT distribution alignment \(\mathcal{L}_{\mathrm{align}}\), and action supervision \(\mathcal{L}_{\mathrm{act}}\). The alignment loss uses a frozen CoT agent to match text- and latent-conditioned reasoning distributions, supervising both the latent agent and observer.}
    \label{fig:method}
    \vspace{-2em}
\end{figure*}

LARVA learns a continuous interface between visual observation and sequential decision making. A reasoner selects which video evidence to acquire, and an observer encodes the selected frames into latent tokens that condition subsequent decisions. We distill this interaction from teacher trajectories using action and answer supervision, together with auxiliary observation reconstruction and CoT-distribution alignment. The central design is to train the observation interface jointly with the reasoner, so that its representations support both interpreting the evidence and decision-making.

\subsection{Sequential Visual Evidence Acquisition}
\label{sec:method_formulation}

Given a video $V$ and a question $Q$, the agent alternates between reasoning, acquisition, and observation. Let $\mathcal{H}_{t-1}$ denote the retained interaction context before decision $t$. The reasoner first computes a latent reasoning block $Z_t$, then generates either a tool call $A_t$ or a final answer. A nonterminal action specifies a tool type, a temporal interval when applicable, and an optional query. Executing it selects frames $X_t$ from $V$; the observer returns a set of latent observation tokens $O_t^i$ for each frame $X_t^i$, paired with the frame's timestamp, which enters the interaction context for subsequent decisions. The previous $Z_t$ is also retained during the next round of decision making. A terminal answer triggers no further observation.

We use $T$ to count nonterminal acquisition rounds and index the terminal answer decision by $T+1$. Thus, the final answer is conditioned on $\mathcal{H}_T$, which includes the observations returned by all completed acquisitions. Tool calls and answers remain explicit text, while intermediate reasoning and observation use continuous states.

\paragraph{Observation tools.}
We use the same VideoSeek~\citep{videoseek} tools for controlled textual and latent students: \textbf{Overview} uniformly samples 64 frames across the video, while \textbf{Skim} and \textbf{Focus} each sample 16 frames from a specified interval for cue search and targeted inspection, respectively. Appendix~\ref{app:method_tools} details their roles.

\subsection{Latent Reasoner--Observer Interface}
\label{sec:method_architecture}

LARVA separates two computational roles: a reasoner $\mathcal{R}_{\theta}$ for action selection and answering, and an observer $\mathcal{O}_{\phi}$ for visual processing. In our implementation, the roles share a backbone and use separate LoRA adapters~\citep{hu2022lora}. The role-specific parameters $\theta$ and $\phi$ include a projector after the backbone's last layer (Appendix~\ref{app:method_decoding}). Raw visual tokens are processed by the observer; the reasoner receives its continuous outputs rather than the frames or a decoded description.

\paragraph{Latent reasoning.}
Before each decision, the reasoner autoregressively computes a fixed-length block $Z_t=[z_t^1,\ldots,z_t^k]$, where $z_t^i\in\mathbb{R}^{d}$ and $d$ is the latent embedding dimension. Writing $f_{\theta}$ for its continuous-state update, we have
\begin{equation}
    z_t^i=f_{\theta}(Q,\mathcal{H}_{t-1},z_t^{<i}),
    \qquad i=1,\ldots,k.
    \label{eq:latent_reasoning}
\end{equation}
Here $z_t^{<i}$ denotes preceding latents in the block. The model then decodes a textual tool call or answer conditioned on $(Q,\mathcal{H}_{t-1},Z_t)$; Appendix~\ref{app:method_decoding} gives the token-level factorization.

\paragraph{Latent observation.}
Let $C_t^{\mathrm{obs}}$ denote the observer input, containing the selected frames $X_t$, tool type $c_t$, and query $q_t$ when present. The observer autoregressively computes $n$ continuous states:
\begin{equation}
    o_t^i=g_{\phi}(C_t^{\mathrm{obs}},o_t^{<i}),
    \qquad O_t=[o_t^1,\ldots,o_t^n]\in\mathbb{R}^{n\times d}.
    \label{eq:latent_observation}
\end{equation}
Here $g_{\phi}$ is the observer's continuous-state update. The resulting $O_t$ is inserted into the reasoner's context without decoding it into words. This makes the observer's output a learned interface to subsequent reasoning and acquisition. Each observation contributes a fixed number of latent positions to the reasoner's context, although the context still grows across rounds and the observer must still process the selected visual input.

\subsection{Trajectory Distillation and Joint Optimization}
\label{sec:method_training}

\paragraph{Textual supervision and initialization.}
A teacher reasoner and a teacher observer interact with the video to generate a trajectory
\begin{equation}
    \widetilde{\mathcal{T}}
    =\left(\{(\widetilde R_t,\widetilde A_t,\widetilde O_t)\}_{t=1}^{T},
    \widetilde y\right),
    \label{eq:teacher_trajectory}
\end{equation}
where $\widetilde R_t$ is textual reasoning, $\widetilde A_t$ is a tool call, $\widetilde O_t$ is the teacher observer's response to the selected frames, and $\widetilde y$ is the teacher's final answer. A terminal reasoning trace, when available, is denoted by $\widetilde R_{T+1}$. We first train a textual CoT student on the teacher reasoning, actions, and answers, and distill a textual observer from the teacher observations. These models initialize the latent reasoner and observer, respectively. A frozen copy of the trained CoT student, denoted by $p_{\bar\psi}$, provides the reference distributions for alignment.

\paragraph{Replaying acquisition trajectories.}
During training, we replay teacher acquisition actions while replacing textual intermediates with generated latent blocks. The reasoner predicts each teacher action, whose execution supplies frames for the latent observer and the next decision. At inference, the reasoner selects its own actions. Appendix~\ref{app:method_replay} details replay and teacher forcing.

\paragraph{Action and answer supervision.}
We supervise explicit decisions through the generated latent context:
\begin{equation}
\begin{aligned}
    \mathcal{L}_{\mathrm{act}}
    ={}&-\sum_{t=1}^{T}\sum_j
    \log p_{\theta}(\widetilde a_{t,j}\mid
    Q,\mathcal{H}_{t-1},Z_t,\widetilde a_{t,<j})
    -\sum_j\log p_{\theta}(\widetilde y_j\mid
    Q,\mathcal{H}_T,Z_{T+1},\widetilde y_{<j}).
\end{aligned}
\label{eq:decision_loss}
\end{equation}
Here $Z_{T+1}$ precedes the terminal answer. Each target token is scored using only its preceding target prefix; serialization and teacher-forcing details are in Appendix~\ref{app:method_replay}.

The continuous observation interface allows downstream decision losses to update the observer as well as the reasoner. In particular, $O_t$ can influence actions at steps $t+1$ and later, and the terminal answer; it cannot influence the already-selected action at step $t$. This supervision connects the representation of acquired evidence to its subsequent use. The teacher actions and frame-selection operations are fixed during replay, so this gradient path runs through latent computation rather than through the discrete acquisition operation.

\paragraph{Observation reconstruction.}
To anchor latent observations to teacher-described visual content, an auxiliary textual decoding branch reconstructs $\widetilde O_t$ from $O_t$. Denoting its decoding distribution by $p_{\mathrm{rec}}$, we minimize
\begin{equation}
    \mathcal{L}_{\mathrm{obs}}
    =-\sum_{t=1}^{T}\sum_j
    \log p_{\mathrm{rec}}(\widetilde o_{t,j}\mid O_t,\widetilde o_{t,<j}).
    \label{eq:observation_reconstruction}
\end{equation}
This objective encourages $O_t$ to preserve information needed to recover the teacher observation. It complements decision supervision, which trains the interface for later actions and answering. Textual reconstruction is an auxiliary training branch; it is not part of the inference loop.

\paragraph{CoT-distribution alignment.}
To supervise how observations combine with prior reasoning, we compare next-CoT distributions under the frozen reference $p_{\bar\psi}$. Let $q_{t,j}$ and $p_{t,j}$ denote its predictions from textual and latent contexts, respectively, at token $j$ after acquisition $t$. Both branches use the same question, teacher actions, and next-CoT prefix; only reasoning and observation representations differ. Appendix~\ref{app:method_alignment} defines the contexts and token distributions explicitly.

Let $\mathcal I$ contain the acquisition steps with an available $\widetilde R_{t+1}$. We align predictions at corresponding token positions using
\begin{equation}
    \mathcal{L}_{\mathrm{align}}
    =\sum_{t\in\mathcal I}\sum_j
    D_{\mathrm{KL}}\!\left(
    \operatorname{sg}[q_{t,j}]\,\Vert\,p_{t,j}\right),
    \label{eq:cot_alignment}
\end{equation}
where $\operatorname{sg}$ denotes stop-gradient. The reference parameters remain frozen, while derivatives through the latent-context branch supervise the states supplied by the reasoner and observer. Alignment thus encourages latent contexts to support next-reasoning predictions similar to those induced by teacher text, without requiring direct targets for the latent vectors.

\paragraph{Overall objective.}
The three losses provide complementary supervision for explicit decisions, observation content, and the interpretation of latent contexts:
\begin{equation}
    \mathcal{L}
    =\lambda_{\mathrm{act}}\mathcal{L}_{\mathrm{act}}
    +\lambda_{\mathrm{obs}}\mathcal{L}_{\mathrm{obs}}
    +\lambda_{\mathrm{align}}\mathcal{L}_{\mathrm{align}}.
    \label{eq:joint_objective}
\end{equation}
The coefficients $\lambda_{\mathrm{act}}$, $\lambda_{\mathrm{obs}}$, and $\lambda_{\mathrm{align}}$ control their relative contributions. Joint optimization couples the observer's representations to the reasoner's use of those representations, while reconstruction and alignment supply supervision from the textual teacher trajectories.

\subsection{Inference}
\label{sec:method_inference}

At inference, the reasoner generates a latent reasoning block and then selects a tool call or produces the final answer. Each tool call acquires frames for the observer, whose latent output conditions the following decision. The loop ends when the reasoner answers or reaches the acquisition limit, after which it answers using the accumulated context. Our main configuration uses 8 latent tokens per reasoning and observation block and permits at most eight acquisition rounds. Neither textual CoT nor textual observations are decoded during inference; tool calls and final answers remain textual.
\vspace{-8pt}
\section{Experiments}
\vspace{-8pt}
\label{sec:experiments}

Our experiments evaluate LARVA as a complete observation--reasoning--action loop. We first test whether it improves video understanding over students distilled from the same teacher trajectories. We then examine the contributions of latent reasoning, latent observations, and joint optimization, before evaluating inference cost and the usefulness of the acquired evidence. Answerer swaps, a human assessment of incremental evidence, and teacher-trajectory comparisons provide complementary views of acquisition behavior.

\subsection{Experimental Setup}
\label{sec:experimental_setup}

\paragraph{Training data and implementation.}
We generate 20,000 interaction trajectories from VideoMarathon~\citep{videomarathon}, the VideoHolmes training set~\citep{videoholmes}, and LongVideo-Reason~\citep{chen2025scaling}, filtering trajectories with incorrect final answers or invalid tool calls. DeepSeek-V4.1-Flash~\citep{deepseekai2026deepseekv41flash} and Qwen3.8-27B~\citep{qwen38} serve as the teacher reasoner and observer. We instantiate LARVA with Qwen2.5-VL-7B~\citep{bai2025qwen25vltechnicalreport} and Qwen3-VL-4B~\citep{bai2025qwen3vltechnicalreport}, sharing a backbone with separate reasoner and observer LoRA adapters. The main configuration uses eight latent tokens per reasoning and observation block and at most eight acquisition rounds. Initialization, optimization settings, and acquisition tools are detailed in Appendix~\ref{app:experimental_setup}.

\paragraph{Benchmarks and comparisons.}
We report accuracy on VideoHolmes~\citep{videoholmes}, Video-MME~\citep{videomme}, LVBench~\citep{lvbench}, LongVideoBench~\citep{wu2024longvideobench}, and MLVU~\citep{MLVU}, excluding Video-MME captions from model input. CoT Policy learns the teacher's reasoning, tool calls, and answers; Direct Policy omits the reasoning traces. Both use the same distilled textual observer and source trajectories, controlling for the supervision data. EVA~\citep{EVA}, VideoMind~\citep{videomind}, DyLaR~\citep{dylar}, and MACF~\citep{macf} provide broader comparisons but differ in architecture and training. Unless otherwise stated, diagnostic experiments use Qwen2.5-VL-7B.

\subsection{Main Results}
\label{sec:main_results}
\begin{table}[t]
\centering
\caption{\textbf{Main results.} Accuracy (\%) on five video-understanding benchmarks. Teacher models are listed for reference. CoT Policy and Direct Policy are controlled students distilled from the same teacher trajectories as LARVA.}
\label{tab:main_results}
\small
\renewcommand{\arraystretch}{1.1}
\setlength{\tabcolsep}{1.4pt}
\begin{tabular}{lccccc}
\toprule
Method & \makecell{VideoHolmes\\{\scriptsize\citep{videoholmes}}} & \makecell{Video-MME\\{\scriptsize\citep{videomme}}} & \makecell{LVBench\\{\scriptsize\citep{lvbench}}} & \makecell{LongVideoBench\\{\scriptsize\citep{wu2024longvideobench}}} & \makecell{MLVU\\{\scriptsize\citep{MLVU}}} \\
\midrule
\grouprow{6}{Teacher models} \\
\makecell[l]{DeepSeek-V4.1-Flash\\{\scriptsize\citep{deepseekai2026deepseekv41flash}}} & 60.4 & 80.1 & 63.3 & 71.9 & 77.2 \\
Qwen3.8-27B~{\scriptsize\citep{qwen38}} & 50.3 & 72.5 & 56.8 & 67.4 & 75.2 \\
\midrule
\grouprow{6}{Qwen2.5-VL-7B~{\scriptsize\citep{bai2025qwen25vltechnicalreport}}} \\
EVA~{\scriptsize\citep{EVA}}             & 37.2 & \underline{60.2} & \underline{43.3} & 55.0 & 68.3 \\
VideoMind~{\scriptsize\citep{videomind}} & \underline{41.0} & 58.2 & 40.8 & 56.3 & 64.4 \\
DyLaR~{\scriptsize\citep{dylar}}         & \underline{41.0} & 59.8 & 41.2 & \underline{57.4} & -- \\
MACF~{\scriptsize\citep{macf}}           & -- & 58.2 & 37.7 & 52.5 & 47.6 \\
CoT Policy    & 39.2 & 59.9 & 41.3 & 56.8 & \underline{68.4} \\
Direct Policy & 34.6 & 53.4 & 37.6 & 52.0 & 61.8 \\
\oursrow LARVA (Ours) & \textbf{44.3} & \textbf{62.4} & \textbf{45.0} & \textbf{58.1} & \textbf{68.8} \\
\midrule
\grouprow{6}{Qwen3-VL-4B~{\scriptsize\citep{bai2025qwen3vltechnicalreport}}} \\
DyLaR~{\scriptsize\citep{dylar}} & \underline{47.5} & 57.6 & 38.5 & \underline{56.8} & -- \\
CoT Policy    & 42.7 & \underline{60.2} & \underline{39.1} & 55.0 & \underline{64.9} \\
Direct Policy & 33.2 & 55.8 & 36.7 & 52.3 & 59.5 \\
\oursrow LARVA (Ours) & \textbf{48.2} & \textbf{62.1} & \textbf{41.3} & \textbf{57.4} & \textbf{66.7} \\
\bottomrule
\end{tabular}
\end{table}

\paragraph{Consistent gains across benchmarks and model scales.}
LARVA outperforms the CoT student on all five benchmarks with both backbones (Table~\ref{tab:main_results}). With Qwen2.5-VL-7B, it improves accuracy from 39.2 to 44.3 on VideoHolmes, 59.9 to 62.4 on Video-MME, and 41.3 to 45.0 on LVBench: gains of \textbf{5.1, 2.5, and 3.7 percentage points}, respectively. It also gains 1.3 points on LongVideoBench and 0.4 on MLVU. With Qwen3-VL-4B, the corresponding gains are 5.5, 1.9, 2.2, 2.4, and 1.8 points. The unweighted mean improvement over the five benchmarks is 2.6 points for 7B and 2.8 points for 4B.

LARVA also achieves the highest reported accuracy within each backbone group in Table~\ref{tab:main_results}. For the 7B backbone, it exceeds the strongest listed alternative by 3.3 points on VideoHolmes, 2.2 on Video-MME, and 1.7 on LVBench. The Direct Policy falls below the CoT student on every benchmark with both backbones, whereas LARVA improves on CoT while replacing textual intermediates with learned latent states. Together, these comparisons support learning intermediate representations for the acquisition loop rather than simply omitting intermediate reasoning.

\subsection{Representation and Training Ablations}
\label{sec:representation_ablation}
\begin{table}[t]
\centering
\caption{\textbf{Contributions of latent reasoning and observation.} Accuracy (\%) of students with textual or latent reasoning and observation interfaces, with (\cmark) or without (\xmark) joint training of the reasoner and observer, on Qwen2.5-VL-7B~\citep{bai2025qwen25vltechnicalreport}. All variants use the same training examples and tools.}
\label{tab:repr_ablation}
\small
\renewcommand{\arraystretch}{1.1}
\setlength{\tabcolsep}{5pt}
\begin{tabular}{llc ccc}
\toprule
\multicolumn{3}{c}{Interface} & \multicolumn{3}{c}{Accuracy (\%)} \\
\cmidrule(r){1-3} \cmidrule(l){4-6}
Reasoning & Observation & Joint & \makecell{VideoHolmes\\{\scriptsize\citep{videoholmes}}} & \makecell{Video-MME\\{\scriptsize\citep{videomme}}} & \makecell{LVBench\\{\scriptsize\citep{lvbench}}} \\
\midrule
Textual & Textual & \xmark & 39.2 & 59.9 & 41.3 \\
Latent  & Textual & \xmark & 42.7 & \underline{61.9} & \underline{43.6} \\
Textual & Latent  & \xmark & 38.6 & 59.3 & 40.7 \\
Textual & Latent  & \cmark & \underline{43.5} & 60.8 & 42.0 \\
Latent  & Latent  & \xmark & 37.4 & 55.2 & 39.7 \\
\oursrow Latent & Latent & \cmark & \textbf{44.3} & \textbf{62.4} & \textbf{45.0} \\
\bottomrule
\end{tabular}
\end{table}
\vspace{-12pt}

\paragraph{Latent reasoning improves the textual-observation student.}
Replacing textual reasoning with latent reasoning while retaining textual observations raises accuracy by 3.5, 2.0, and 2.3 points on VideoHolmes, Video-MME, and LVBench (Table~\ref{tab:repr_ablation}). Full LARVA further improves on this variant by 1.6, 0.5, and 1.4 points. The strongest configuration therefore combines latent reasoning with jointly learned latent observations, rather than obtaining its full benefit from latent reasoning alone.

\paragraph{Joint optimization makes the latent interface effective.}
\label{sec:joint_training_ablation}
Latent observations are not sufficient by themselves: without joint training, the textual-reasoning/latent-observation variant scores 38.6, 59.3, and 40.7, below the fully textual student on all three benchmarks. Joint training raises these scores to 43.5, 60.8, and 42.0. The effect is larger when both interfaces are latent: joint optimization improves accuracy from 37.4 to 44.3, 55.2 to 62.4, and 39.7 to 45.0, gaining \textbf{6.9, 7.2, and 5.3-point}.

For the non-joint latent/latent variant, we first train the latent reasoner with textual observations, then freeze it and train the latent observer with the objectives in Section~\ref{sec:method}. Its weaker performance highlights the challenge of fitting a continuous observation interface to a fixed receiver trained on text. Joint training instead allows the reasoner and observer to adapt together, turning the latent/latent configuration from below the textual baseline into the strongest evaluated variant.

\paragraph{Latent-token capacity.}
Increasing observation or reasoning latents from one to eight generally improves accuracy, with smaller gains between six and eight. Very short observation sequences are particularly restrictive, supporting sufficient capacity for visual communication. Appendix~\ref{app:latent_capacity} provides the complete sweep and analysis (Figure~\ref{fig:lat_tok_count}).

\subsection{Inference Cost}
\label{sec:inference_cost}
\begin{table}[t]
\centering
\caption{\textbf{Inference cost} per question, averaged over VideoHolmes, Video-MME, and LVBench. 
Reasoning tokens are generated by the reasoner, and observation tokens are received by it. Latency is measured on NVIDIA H200 GPUs. All local models use 7B backbones; EVA and the students use Qwen2.5-VL-7B~\citep{bai2025qwen25vltechnicalreport}. $^\dagger$Relies on closed-source APIs, so latency includes API response time; excluded from Table~\ref{tab:main_results}.}
\label{tab:efficiency}
\small
\renewcommand{\arraystretch}{1.1}
\setlength{\tabcolsep}{5pt}
\begin{tabular}{lccccc}
\toprule
 & & \multicolumn{2}{c}{Tokens} & & \\
\cmidrule(lr){3-4}
Method & Turns & Reasoning & Observation & Frames & Latency (s) \\
\midrule
EVA~{\scriptsize\citep{EVA}}                      & 2.6 & 361.2  & 13984.4 & 71.4  & \underline{11.8} \\
VideoMind~{\scriptsize\citep{videomind}}          & 3.0 & 16.6   & 19253.2 & 511.5 & 21.4 \\
LongVideoAgent$^\dagger$~{\scriptsize\citep{liu-etal-2026-longvideoagent}} & 2.5 & 3196.9 & 168.8 & 21.7 & 30.8 \\
\midrule
CoT Policy    & 4.8 & 1752.5 & 1457.2 & 112.5 & 48.3 \\
Direct Policy & 4.4 & 0.0    & 1506.5 & 110.9 & 39.0 \\
\oursrow LARVA (Ours) & 3.2 & 12.8 & 577.4 & 75.9 & \textbf{8.2} \\
\bottomrule
\end{tabular}
\end{table}
\vspace{-12pt}

Table~\ref{tab:efficiency} summarizes average inference cost over VideoHolmes, Video-MME, and LVBench. LARVA records 8.2 seconds per question compared with 48.3 for CoT Policy and 39.0 for Direct Policy, corresponding to \textbf{5.9$\times$ and 4.8$\times$ speedups} in the reported setting. Relative to CoT, average acquisition turns decrease from 4.8 to 3.2 and processed frames from 112.5 to 75.9, reductions of 33.3\% and 32.5\%, respectively. Thus, its higher accuracy is accompanied by less visual processing on average.

LARVA also reduces the reported intermediate reasoning positions from 1752.5 to 12.8 and observation positions received by the reasoner from 1457.2 to 577.4. These counts describe the reasoner's interface; the observer still incurs the cost of encoding selected frames. LARVA has the lowest reported latency among the listed systems, but frame counts, token types, architectures, and API overhead differ across systems. We therefore interpret these measurements as the cost of each complete configuration rather than an isolated estimate of the savings from latent representations.

\subsection{Evidence Acquisition Analysis}
\label{sec:evidence_analysis}

\paragraph{Acquired evidence benefits multiple terminal answerers.}
\begin{table}[t]
\centering
\caption{\textbf{Evidence quality via answerer swaps.} Trajectories are generated by the CoT or latent student; at the terminal decision, the acquired evidence is passed to the listed answerer.}
\label{tab:evidence_quality}
\small
\renewcommand{\arraystretch}{1.1}
\setlength{\tabcolsep}{8pt}
\begin{tabular}{ll ccc}
\toprule
Trajectory & Answerer & VideoHolmes & Video-MME & LVBench \\
\midrule
\multirow{3}{*}{CoT} & CoT          & 39.2 & 59.9 & 41.3 \\
                     & Latent       & 42.8 & 60.0 & 41.5 \\
                     & 27B Visual   & 49.9 & 63.7 & 49.6 \\
\midrule
\multirow{3}{*}{Latent} & CoT       & 43.6 & 61.1 & 41.5 \\
 & \ourscell Latent & \ourscell 44.3 & \ourscell 62.4 & \ourscell 45.0 \\
 & 27B Visual       & \textbf{51.6} & \textbf{67.9} & \textbf{52.2} \\
\bottomrule
\end{tabular}
\end{table}

To assess whether gains extend beyond LARVA's own answer decoder, we replace the answerer when the acquisition policy stops or exhausts its action budget. Table~\ref{tab:evidence_quality} compares CoT and LARVA trajectories with each terminal answerer held fixed. LARVA trajectories improve accuracy in all nine answerer--benchmark comparisons. With the CoT answerer, the gains are 4.4, 1.2, and 0.2 points; with the latent answerer, they are 1.5, 2.4, and 3.5 points. A fixed 27B visual answerer gains \textbf{1.7, 4.2, and 2.6 points}, respectively.

The benefit to a separate visual answerer supports the usefulness of LARVA's selected evidence beyond compatibility with its own latent answerer. This complements the end-to-end results: improvements persist when terminal answering is held fixed. The swap experiment evaluates complete acquisition trajectories, whose realized lengths and frame counts may differ; the human study below targets incremental evidence under a fixed additional-frame budget.

\paragraph{Human evaluation of incremental evidence.}
\label{sec:human_evaluation}
We conduct a paired comparison between full LARVA and the latent-reasoning/textual-observation ablation, both using Qwen2.5-VL-7B, on 100 held-out questions. Each model receives the same 64-frame Overview and makes one additional Skim/Focus call returning 16 frames. This fixes the initial visual coverage and the additional-frame budget, focusing the comparison on the usefulness of the next acquisition.

Two independent judges assess anonymized, randomly ordered acquisition sets for new answer-relevant clues absent from the Overview. Model identities, reasoning traces, and answers are hidden; repeated clues do not count. Judgments are Yes, No, or Uncertain, and disagreements are adjudicated. Appendix~\ref{app:human_protocol} details the reference checklist and annotation procedure.

\begin{table}[t]
\centering
\small
\caption{\textbf{Human assessment of incremental evidence.} Evidence-gain rate is adjudicated Yes divided by 100 questions. Uncertain cases remain in the denominator and are reported as counts. Both models receive the same 64-frame Overview and a 16-frame additional acquisition budget.}
\label{tab:human_evidence}
\begin{tabular}{lcc}
\toprule
Model & Evidence-gain rate (\%) & Uncertain (count) \\
\midrule
Text reasoning + latent observation & 11 & 7 \\
Latent reasoning + textual observation & 15 & 5 \\
Full LARVA                            & 35 & 8 \\
\bottomrule
\end{tabular}
\end{table}
\vspace{-8pt}

Full LARVA yields new answer-relevant evidence on \textbf{35\%} of questions, compared with \textbf{15\%} for the latent-reasoning/textual-observation and \textbf{11\%} for text reasoning and latent observation. Full LARVA produces up to \textbf{24-percentage-point gain} under the same additional-frame budget (Table~\ref{tab:human_evidence}). Uncertain cases are retained in the denominator for both models. This result complements the answerer swaps by showing that the full model more often selects frames containing an additional useful clue, independently of its final-answer generation. This measures the first acquisition after a shared Overview; it does not measure evidence coverage across an entire trajectory.

\paragraph{Closer agreement with teacher acquisition trajectories.}
LARVA exceeds both textual students in tool-sequence similarity, temporal IoU, and optional-tool agreement across all three diagnostic benchmarks. Relative to CoT, its tool-sequence scores increase by 4.6, 0.8, and 1.3 points, and temporal IoU by 1.2, 2.3, and 2.0 points. These results support better transfer of teacher acquisition behavior, complementing the answerer swaps that assess the usefulness of acquired evidence. Appendix~\ref{app:trajectory_similarity} gives the metric definitions and complete results (Table~\ref{tab:traj_similarity}).

\vspace{-8pt}
\section{Conclusion}
\vspace{-8pt}
We introduce LARVA, a framework that jointly learns latent reasoning and latent visual observations for agentic video understanding. By combining action supervision, observation reconstruction, and CoT distribution alignment, LARVA learns compact intermediate representations that support multi-round evidence acquisition and answer generation. Experiments across three benchmarks show that LARVA outperforms existing methods while reducing inference latency. Further analyses demonstrate stronger evidence-gathering capability than textual distilled students and closer agreement with the teacher's acquisition behavior. These results highlight joint latent training as an effective approach to building accurate and efficient video agents.
\vspace{-8pt}
\section{Limitations}
\vspace{-8pt}
LARVA has several limitations. First, it operates on sampled visual frames and does not process audio. It therefore cannot directly use dialogue, environmental sounds, or other acoustic cues, limiting its applicability to questions requiring audiovisual understanding.

Second, training relies on teacher-generated trajectories and a frozen CoT student for distribution alignment. These supervision sources may contain errors or suboptimal acquisition decisions that LARVA can inherit. Moreover, training on teacher trajectories may provide limited coverage of the histories encountered during student execution, particularly after an incorrect acquisition.

Finally, compact latent representations are less directly interpretable than textual reasoning and observations. Although tool calls and acquired frames remain inspectable, identifying whether a failure originates in visual compression, reasoning, or evidence selection is more difficult. Compression may also discard fine-grained details needed for some questions.

\vspace{-8pt}
\section*{AI Use Statement}
\vspace{-8pt}
We used generative AI tools to assist with manuscript editing, figure preparation, and discussions of research framing and experimental design. We also used teacher models to generate the training trajectories described in the paper. The authors reviewed AI-assisted content for accuracy and consistency and take responsibility for the manuscript, methodology, and reported results.
\vspace{-8pt}
\section*{Ethics Statement}
\vspace{-8pt}
This work studies efficient evidence acquisition for video understanding using existing research datasets. Video agents may produce incorrect interpretations, inherit biases from their training data and teacher models, or overlook relevant visual evidence. Their outputs therefore require appropriate human oversight in consequential applications. Any use or release of video data and derived training trajectories should respect applicable dataset licenses, privacy requirements, and usage restrictions.
\vspace{-8pt}
\section*{Reproducibility Statement}
\vspace{-8pt}
The paper describes LARVA's architecture, training objectives, teacher-trajectory generation, and evaluation protocols. To support reproducibility, we will provide implementation details, model configurations, prompts, hyperparameters, and evaluation scripts, including the protocols for trajectory similarity, answerer-swap evaluation, and inference-time measurement. We plan to release code and permitted training artifacts, subject to the licenses and usage restrictions of the underlying datasets and models.
\bibliography{iclr2027_conference}
\bibliographystyle{iclr2027_conference}

\appendix
\section{Appendix}

\label{app:method_details}
\label{app:experiments}
This appendix supplements Sections~\ref{sec:method} and~\ref{sec:experiments} with implementation and training details, observation tools, action-decoding and alignment formulations, the complete human-evaluation protocol, and additional analyses of latent-token capacity and teacher-trajectory similarity.

\subsection{Observation Tools}
\label{app:method_tools}
We adopt the \textbf{Overview}, \textbf{Skim}, and \textbf{Focus} interface of VideoSeek~\citep{videoseek}, and use the same tools for our controlled textual and latent students. Overview uniformly samples 64 frames across the full video. Skim samples 16 frames from a reasoner-specified interval and asks the observer to search for query-relevant cues. Focus also samples 16 frames from a specified interval, with a query targeting a concrete sub-question. These operations support global exploration and progressively targeted inspection while keeping acquisition actions externally executable.

\subsection{Projection and Explicit Action Decoding}
\label{app:method_decoding}
We use $\theta$ and $\phi$ to denote the role-specific trainable parameters, including the final projector following the backbone's last-layer output, which we implement as one layer-norm followed by a linear layer.

The prefix $z_t^{<i}$ contains the preceding latent states in the current block. After completing the block, the model switches to textual decoding to predict the next tool call or answer. For an acquisition action,
\begin{equation}
    p_{\theta}(A_t\mid Q,\mathcal{H}_{t-1},Z_t)
    =\prod_j p_{\theta}(a_{t,j}\mid Q,\mathcal{H}_{t-1},Z_t,a_{t,<j}),
    \label{eq:action_prediction}
\end{equation}
where $a_{t,j}$ is the $j$-th token in the serialized tool call.

\subsection{Trajectory Replay and Teacher Forcing}
\label{app:method_replay}
During latent training, we retain the teacher's acquisition actions but replace textual intermediates with generated latent blocks. At step $t$, the reasoner produces $Z_t$ from the current latent context and predicts the teacher action $\widetilde A_t$. We then append the teacher-forced action, execute its frame-sampling operation, and compute $O_t$ from the selected frames. The action and latent observation become available to the next decision. The teacher thus fixes which evidence is acquired during training, while the students learn how to represent and use that evidence. At inference, acquisition actions are generated by the reasoner itself.

Here $Z_{T+1}$ is the reasoning block preceding the terminal answer, and the answer sequence includes the model's terminal response formatting. Each target token is scored using only its preceding target prefix. Teacher forcing therefore supplies the history for subsequent prediction positions without exposing the token being predicted.

\subsection{Alignment Contexts and Token Distributions}
\label{app:method_alignment}
Reconstructing an observation does not directly supervise how it combines with prior reasoning to support the next decision. We therefore compare the next-CoT predictions induced by textual and latent contexts under the same frozen reference $p_{\bar\psi}$. After acquisition $t$, the textual context $C_t^{\mathrm{text}}$ contains teacher reasoning and observations. The corresponding context $C_t^{\mathrm{lat}}$ replaces these fields with generated reasoning and observation latents. The question, teacher actions, and shared contextual fields are held fixed.

For every step with an available next reasoning trace, both branches receive the same teacher-forced prefix $\widetilde R_{t+1,<j}$. Their distributions over the text vocabulary are
\begin{equation}
\begin{aligned}
    q_{t,j}(v)&=p_{\bar\psi}(v\mid C_t^{\mathrm{text}},\widetilde R_{t+1,<j}),\\
    p_{t,j}(v)&=p_{\bar\psi}(v\mid C_t^{\mathrm{lat}},\widetilde R_{t+1,<j}).
\end{aligned}
\label{eq:alignment_distributions}
\end{equation}
The distributions in Equation~\ref{eq:alignment_distributions} are compared at corresponding token positions by the KL objective in Equation~\ref{eq:cot_alignment}. The reference parameters remain frozen, while the latent-context branch retains derivatives to the states generated by the reasoner and observer.

\subsection{Training and Evaluation Setup}
\label{app:experimental_setup}
\paragraph{Training data and implementation.}
We generate 20,000 interaction trajectories from VideoMarathon~\citep{videomarathon}, the VideoHolmes training set~\citep{videoholmes}, and LongVideo-Reason~\citep{chen2025scaling}. DeepSeek-V4.1-Flash~\citep{deepseekai2026deepseekv41flash} serves as the teacher reasoner, producing reasoning traces, tool calls, and final answers; Qwen3.8-27B~\citep{qwen38} serves as the teacher observer, returning textual observations of the requested frames. Teacher trajectories with an incorrect final answer or at least one invalid tool call are filtered out. These trajectories supervise explicit decisions and provide textual targets for learning the latent interface.

We instantiate LARVA with Qwen2.5-VL-7B~\citep{bai2025qwen25vltechnicalreport} and Qwen3-VL-4B~\citep{bai2025qwen3vltechnicalreport}. The reasoner and observer share a backbone with separate LoRA adapters ($r=16$, $\alpha=32$), initialized from the distilled CoT student and textual observer, respectively. We train for one epoch with a learning rate of $10^{-4}$ and set $\lambda_{\mathrm{act}}=\lambda_{\mathrm{obs}}=\lambda_{\mathrm{align}}=1$. The main configuration uses eight latent tokens per reasoning and observation block and permits at most eight acquisition rounds. Tool calls and final answers are decoded as text. The acquisition tools are Overview (64 frames) and Skim/Focus (16 frames each), as defined in Section~\ref{sec:method_formulation}. Since we train LARVA from already trained CoT and captioner models, the CoT policy, Direct policy, and captioner in Section~\ref{app:experiments} are trained for two epochs instead of one to ensure fair comparison.

\subsection{Human-Evaluation Protocol}
\label{app:human_protocol}
We conduct a paired comparison between full LARVA and the latent-reasoning/textual-observation ablation, both using Qwen2.5-VL-7B, on 100 held-out questions. Each model receives the same 64-frame Overview and makes one additional Skim/Focus call returning 16 frames. This fixes the initial visual coverage and the additional-frame budget, focusing the comparison on the usefulness of the next acquisition.

Two independent judges inspect the question, a reference evidence checklist, the shared Overview, and anonymized acquisition sets presented in randomized order. Model identities, reasoning traces, and answers are hidden, and the checklist is not provided to either model. Judges label whether the actual sampled frames reveal a new answer-relevant clue absent from the Overview as Yes, No, or Uncertain, with a brief justification. Repeated clues do not count as new evidence. Disagreements are adjudicated.

Evidence-gain rate is the number of adjudicated Yes judgments divided by all 100 questions; Uncertain cases remain in the denominator and are reported separately in Table~\ref{tab:human_evidence}. This evaluation measures the first acquisition after a shared Overview, rather than evidence coverage across an entire trajectory.

\subsection{Latent-Token Capacity}
\label{app:latent_capacity}
\label{sec:observation_scaling}

\begin{figure*}[t]
    \centering
    \includegraphics[width=0.9\linewidth]{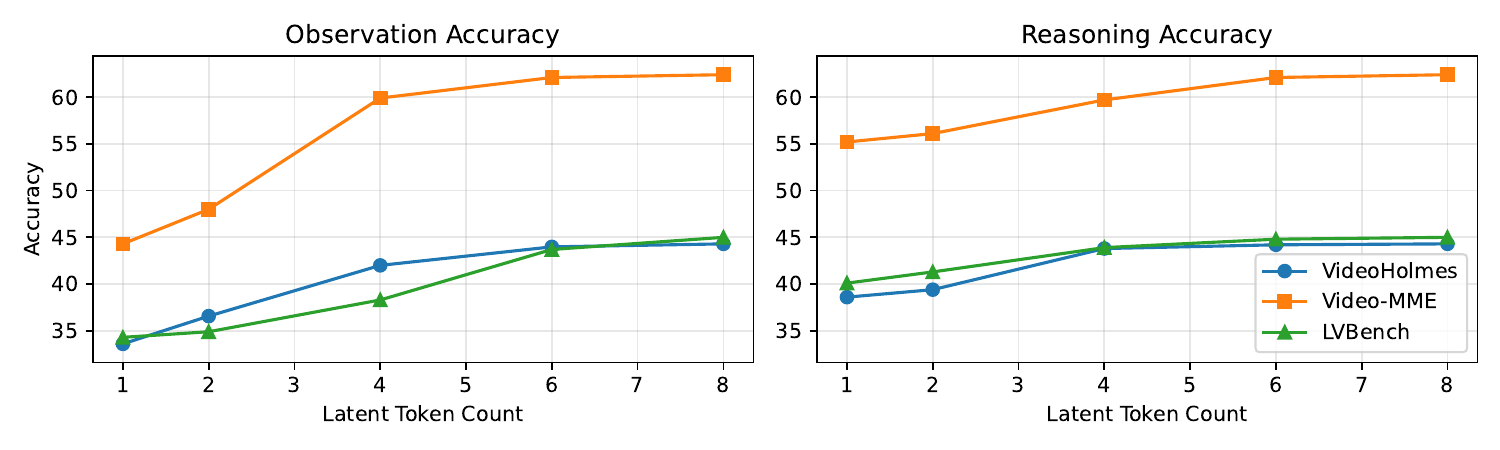}
    \caption{\textbf{Left:} Accuracy vs. latent observation token count. \textbf{Right:} Accuracy vs. latent reasoning token count. The performance increases with both hyperparameters and plateaus at around 8 tokens. }
    \label{fig:lat_tok_count}
\end{figure*}
Figure~\ref{fig:lat_tok_count} varies the number of observation or reasoning latents from one to eight. Accuracy generally improves with capacity on all three benchmarks, with smaller changes between six and eight tokens. The observation sweep is particularly sensitive to very short sequences, consistent with an overly narrow interface restricting the visual information available for downstream decisions. Eight tokens provide a strong accuracy setting within the evaluated range. The figure measures accuracy, so it does not by itself establish an optimal latency--accuracy trade-off.

\subsection{Teacher-Trajectory Similarity}
\label{app:trajectory_similarity}
\begin{table}[t]
\centering
\caption{\textbf{Similarity to teacher trajectories.} Answer accuracy (\%) and agreement between student and teacher acquisition trajectories on Qwen2.5-VL-7B. For each benchmark, the best result is in \textbf{bold}.}
\label{tab:traj_similarity}
\small
\renewcommand{\arraystretch}{1.1}
\setlength{\tabcolsep}{5.5pt}
\begin{tabular}{ll c ccc}
\toprule
 & & & \multicolumn{3}{c}{Similarity to teacher trajectory} \\
\cmidrule(l){4-6}
Benchmark & Method & Accuracy & \makecell{Tool\\Sequence} & \makecell{Temporal\\IoU} & \makecell{Optional\\Tools}\\
\midrule
\multirow{3}{*}{VideoHolmes} & Direct & 34.6 & 54.6 & 36.2 &26.1\\
                             & CoT    & 39.2 & 57.3 & 38.2 & 29.5 \\
 & \ourscell LARVA & \ourscell\textbf{44.3} & \ourscell\textbf{61.9} & \ourscell\textbf{39.4} & \ourscell \textbf{30.7} \\
\midrule
\multirow{3}{*}{Video-MME}   & Direct & 53.4 & 60.6 & 30.4 & 29.0  \\
                             & CoT    & 59.9 & 63.6 & 30.5 & 35.9   \\
 & \ourscell LARVA & \ourscell\textbf{62.4} & \ourscell\textbf{64.4} & \ourscell\textbf{32.8} & \ourscell \textbf{40.9} \\
\midrule
\multirow{3}{*}{LVBench}     & Direct & 37.6 & 61.5 & 30.1 & 33.3 \\
                             & CoT    & 41.3 & 63.6 & 30.6 & 34.2  \\
 & \ourscell LARVA & \ourscell\textbf{45.0} & \ourscell\textbf{64.9} & \ourscell\textbf{32.6} & \ourscell \textbf{46.1} \\
\bottomrule
\end{tabular}
\end{table}

Table~\ref{tab:traj_similarity} compares student trajectories with the teacher using three metrics. \textbf{Tool Sequence} is normalized Levenshtein similarity, $1-d_{\mathrm{Lev}}(s,t)/\max(|s|,|t|)$, where $s$ and $t$ are student and teacher tool-name sequences. \textbf{Temporal IoU} is $\mu(A_s\cap A_t)/\mu(A_s\cup A_t)$, where $A_s$ and $A_t$ are the unions of their retrieved temporal intervals and $\mu$ denotes duration. Following \citet{yang2026agentslooksamequantifying}, \textbf{Optional Tool Agreement} measures agreement on using or omitting tool types that appear in some but not all successful reference trajectories for the same question. LARVA outperforms both textual students on all three metrics across all benchmarks. In benchmark order, gains over CoT are 4.6/0.8/1.3 points for Tool Sequence, 1.2/2.3/2.0 for Temporal IoU, and 1.2/5.0/11.9 for Optional Tool Agreement. These results support better transfer of teacher acquisition behavior; the answerer-swap evaluation separately assesses the utility of the acquired evidence.

\end{document}